%% file: main.tex
\documentclass[]{fairmeta}

\input{preamble}

\usepackage{tcolorbox}
\usepackage{amsmath}
\usepackage{amssymb}
\usepackage{xspace}

\title{\algname: Unifying Representation and Generation with Contrastive-Autoregressive Finetuning}

\author[1,2,\dagger]{Hao Yu}
\author[1]{Zhuokai Zhao}
\author[1]{Shen Yan}
\author[1]{Lukasz Korycki}
\author[1]{Jianyu Wang}
\author[1]{Baosheng He}
\author[1]{Jiayi Liu}
\author[1]{Lizhu Zhang}
\author[1]{Xiangjun Fan}
\author[1]{Hanchao Yu}

\affiliation[1]{Meta}
\affiliation[2]{Boston University}
\contribution[\dagger]{Work done during an internship at Meta.}

\abstract{
The rapid advancement of large vision-language models (LVLMs) has driven significant progress in multimodal tasks, enabling models to interpret, reason, and generate outputs across both visual and textual domains. 
While excelling in generative tasks, existing LVLMs often face limitations in tasks requiring high-fidelity representation learning, such as generating image or text embeddings for retrieval. 
Recent work has proposed finetuning LVLMs for representational learning, but the finetuned model often loses its generative capabilities due to the representational learning training paradigm.
To address this trade-off, we introduce \textbf{\algname}, a \textbf{c}ontrastive-\textbf{a}utoregressive \textbf{f}in\textbf{e}-tuning framework that enhances LVLMs for both representation and generative tasks.
By integrating a contrastive objective with autoregressive language modeling, our approach unifies these traditionally separate tasks, achieving state-of-the-art results in both multimodal retrieval and multimodal generative benchmarks, including
object hallucination (OH) mitigation. 
\algname establishes a novel
framework that synergizes embedding and generative functionalities in a single model, setting a foundation for future multimodal models that excel in both retrieval precision and coherent output generation.
}

\date{\today}
\correspondence{Hanchao Yu at \email{hanchaoyu@meta.com}}

\metadata[Code]{\url{https://github.com/haoyu-bu/CAFe}}
\metadata[Project Page]{\url{https://haoyu-bu.github.io/CAFeUnify/}}

\begin{document}

\maketitle

\input{sec/1_intro}
\input{sec/2_relatedwork}
\input{sec/3_method}
\input{sec/4_experiments}
\input{sec/5_ablation_analysis}
\input{sec/6_conclusion}

\clearpage
\newpage
\bibliographystyle{assets/plainnat}
\bibliography{main}

\clearpage
\newpage
\beginappendix
\input{sec/X_appendix}

\end{document}

%% file: preamble.tex
\newcommand{\figref}[1]{Fig.~\ref{#1}}
\newcommand{\tabref}[1]{Table~\ref{#1}}
\newcommand{\eqnref}[1]{\text{Eq.}~(\ref{#1})}
\newcommand{\secref}[1]{\S\ref{#1}}

\newcommand{\algname}{\textsc{CAFe}\xspace}

%% file: sec/1_intro.tex
\section{Introduction}\label{sec:introduction}

\begin{figure*}[t]
\includegraphics[width=\textwidth]{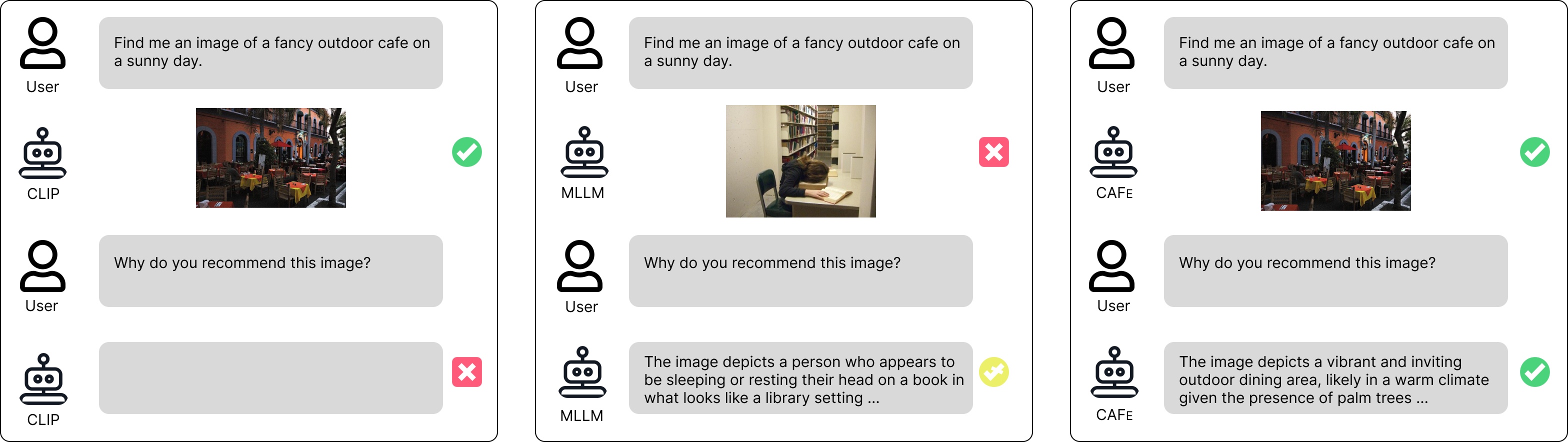}
\captionof{figure}{Capabilities of various vision-language models. While encoder-based models, e.g., CLIP, excel in generating vision-text aligned embeddings and show promising results in image-text retrieval, they fall short in producing free-form text and reasoning about retrieved images (left). Conversely, Multimodal Large Language Models (MLLMs) have shown remarkable success in multimodal understanding and generation, but their direct embeddings yield suboptimal retrieval results (middle). 
\algname effectively bridges this gap by integrating representation learning and language generation, enabling not only retrieval but also advanced generative capabilities (right). 
}
\label{fig:teaser}
\end{figure*}

The rapid development of large vision-language models (LVLMs) has unlocked new capabilities in multimodal understanding, enabling models to perceive, interpret, and reason over complex information across both visual and textual inputs~\citep{du2022survey, yin2023survey, ghosh2024exploring, chen2024autoprm, zhang2024vision, chen2024mj}.
Typically, an LVLM comprises a modality-specific encoder, such as a vision transformer~\citep{dosovitskiy2020image,radford2021learning}, a pre-trained decoder-only large language model (LLM), such as LLaMA~\citep{touvron2023llama,dubey2024llama}, and a bridging projector that facilitates vision-language alignment between the encoder and the decoder.
Popular LVLMs such as LLaVA~\citep{liu2023improvedllava, liu2024visual, li2024llava}, InternVL~\citep{chen2023internvl, gao2024mini}, and BLIP~\citep{li2022blip, li2023blip} have demonstrated significant improvements in tasks such as visual question answering (VQA)~\citep{wu2017visual, de2023visual} by aligning both visual and textual modalities during generations.

Despite the success in generative tasks, most current LVLMs exhibit limitations in tasks that require strong representation learning, such as unimodal and cross-modal retrieval~\citep{jiang2024vlm2vec}.
These tasks demand high-quality embeddings that capture nuanced, context-rich information across modalities, which are essential for precision and robustness. 
However, this remains a challenge for decoder-only architectures designed primarily for sequential output generation, as they are inherently less suited to expressive representation generation. 
This structural limitation means that while these LVLMs perform well in generative tasks, their retrieval performance, where high-fidelity representations are critical, remains suboptimal.

Nonetheless, the open-world knowledge embedded within pre-trained LLMs provides a strong foundation for generating effective embeddings, especially with additional contrastive fine-tuning~\citep{behnamghader2024llm2vec, jiang2024vlm2vec}.
However, modifications such as switching causal attention to bi-directional~\citep{behnamghader2024llm2vec} limits their generative capabilities, which greatly narrows down their downstream use cases and presents a trade-off between representation learning and open-ended generation rather than a unified gain.
To this end, we propose a novel \textbf{c}ontrastive-\textbf{a}utoregressive \textbf{f}in\textbf{e}-tuning (\textbf{\algname}) framework that enhances the capabilities of LVLMs, allowing them to produce high-quality embeddings suitable for retrieval tasks without compromising generative functionality. 
By incorporating contrastive loss during fine-tuning, \algname aligns multimodal representations across text and image pairs, strengthening the model’s potential for both accurate retrieval and coherent generation. 
Through extensive experiments, we demonstrate that the proposed framework achieves state-of-the-art performance in multimodal retrieval, leveraging MLLM's advanced multimodal understanding capabilities. 
Additionally, the modality gap~\citep{liang2022mind} in 
the multimodal representation space is removed, potentially improving cross-modal consistency and enhancing multimodal performance.
On the other hand, the model also shows strong generative abilities and robustness against object hallucinations (OH)~\citep{biten2022let,zhou2023analyzing, chen2024halc, fang2024uncertainty}, a common issue in modern LVLMs, potentially due to the incorporation of the contrastive objective, which enhances the model's ability to differentiate fine-grained details.

As shown in \figref{fig:teaser}, traditional encoder-based models like CLIP~\citep{radford2021learning} and its variants such as RankCLIP~\citep{zhang2024rankclip} excel in creating aligned embeddings for image-text retrieval but lack in generative reasoning.
In contrast, our proposed \algname, depicted on the right side of \figref{fig:teaser}, successfully integrates representation learning with generative capabilities.
In other words, \algname establishes a unified framework that simultaneously supports generative and retrieval capabilities, bridging the gap between these traditionally distinct domains.
And our contributions can be summarized as follow:
\begin{itemize}
    \item We introduce a unified contrastive-autoregressive fine-tuning framework that enables LVLMs to excel in both retrieval and generation tasks.
    \item Evaluated across multimodal benchmarks, the model achieves state-of-the-art (SOTA) results in multimodal retrieval, while also demonstrating strong multimodal understanding capability and robustness against object hallucinations (OH), which significantly enhances the versatility and performance of LVLMs across complex multimodal tasks.
    \item To the best of our knowledge, \algname is the first unified contrastive-autoregressive fine-tuning framework for LVLMs,
    providing additional insight for developing better-rounded multimodal models.
\end{itemize}

%% file: sec/2_relatedwork.tex
\section{Related Work}\label{sec:related_work}

\subsection{Multimodal Large Language Models}
Recent advances in multimodal large language models (MLLMs) have led to the development of models that can receive, reason, and output with multimodal information. 
Notable examples include BLIP-2~\citep{li2023blip}, Flamingo~\citep{alayrac2022flamingo}, and LLaVA~\citep{liu2024visual}, which integrate additional encoders into textual LLMs to process inputs from other modalities. 
A typical MLLM architecture consists of an LLM, a modality encoder, and a projector that connects them. 
For example, LLaVA~\citep{liu2024visual} leverages a pretrained LLM and a pretrained vision encoder, followed by a two-stage training process. 
In the first stage, the model aligns text and image inputs using image-text pairs, training only the projector between the LLM and the modality encoder. 
The second stage involves fine-tuning the model on a visual instruction tuning dataset for improved instruction following ability.

State-of-the-art MLLMs have demonstrated excellent performance in various vision scenarios, including single-image~\citep{dai2023instructblip,zhu2023minigpt,zhang2024llamaadapter}, multi-image~\citep{jiang2405mantis,li2023empowering,li2023fine}, and video settings~\citep{li2025llama,lin2023video, zhang2024beyond}. 
Furthermore, models like LLaVA-OneVision~\citep{li2024llava} have achieved state-of-the-art results across a broad range of tasks and in all three vision scenarios~\citep{li2024llavanext,zhang2024internlm}. 
Despite these success, the use of MLLMs for enhancing text and image understanding in multimodal representations remains relatively under-explored. 
%
To address this gap, we leverage a MLLM for unified representation learning and language generation by introducing a contrastive-autoregressive fine-tuning framework.

\subsection{Vision-language Representation Learning}
Recent vision-language models (VLMs) have achieved remarkable success by pretraining on large-scale image-text pairs, yielding strong visual and textual representations that enable zero-shot adaptation to tasks such as image classification and cross-modal retrieval~\citep{zhang2024rankclip, zhang2024vision, ghosh2024exploring}. 
One of the pioneering works in this area is Contrastive Language-Image Pretraining (CLIP)~\citep{radford2021learning}, which introduced a dual-encoder architecture pretrained with contrastive objectives on noisy image-text pairs. 
Subsequent works have built upon its foundation by refining data strategies or proposing alternative learning objectives~\citep{li2022blip,zhai2023sigmoid,li2023blip,cherti2023reproducible, zhao2024multimodal, zhang2024rankclip}. 

One line of research combines contrastive learning with language modeling~\citep{yu2022coca, li2023blip}. 
For example, CoCa~\citep{yu2022coca} trains an image-text encoder-decoder model jointly with contrastive loss and captioning loss. 
BLIP~\citep{li2023blip} optimizes a combination of understanding-based and generation-based objectives during pretraining. 
These methods have achieved promising performance and demonstrated flexible transferability to both vision-language understanding and generation tasks.

With the advancement of MLLMs, recent studies have explored their application in multimodal representation learning~\citep{zhang2024notellm,jiang2024e5,jiang2024vlm2vec}.
For example, NoteLLM-2~\citep{zhang2024notellm} leverages MLLMs to enhance multimodal representations for item-to-item (I2I) recommendations. 
Concurrently with our work, VLM2VEC~\citep{jiang2024vlm2vec} presents a multimodal embedding benchmark and adapts an MLLM for contrastive representation learning. 
However, they have not investigated how this adaptation affects the language generation capabilities of MLLMs. 
Our work addresses this gap by jointly modeling both embedding and language generation tasks based on a state-of-the-art MLLM.

%% file: sec/3_method.tex
\section{Method}\label{sec:method}
\begin{figure*}[t]
    \centering
    \includegraphics[width=\textwidth]{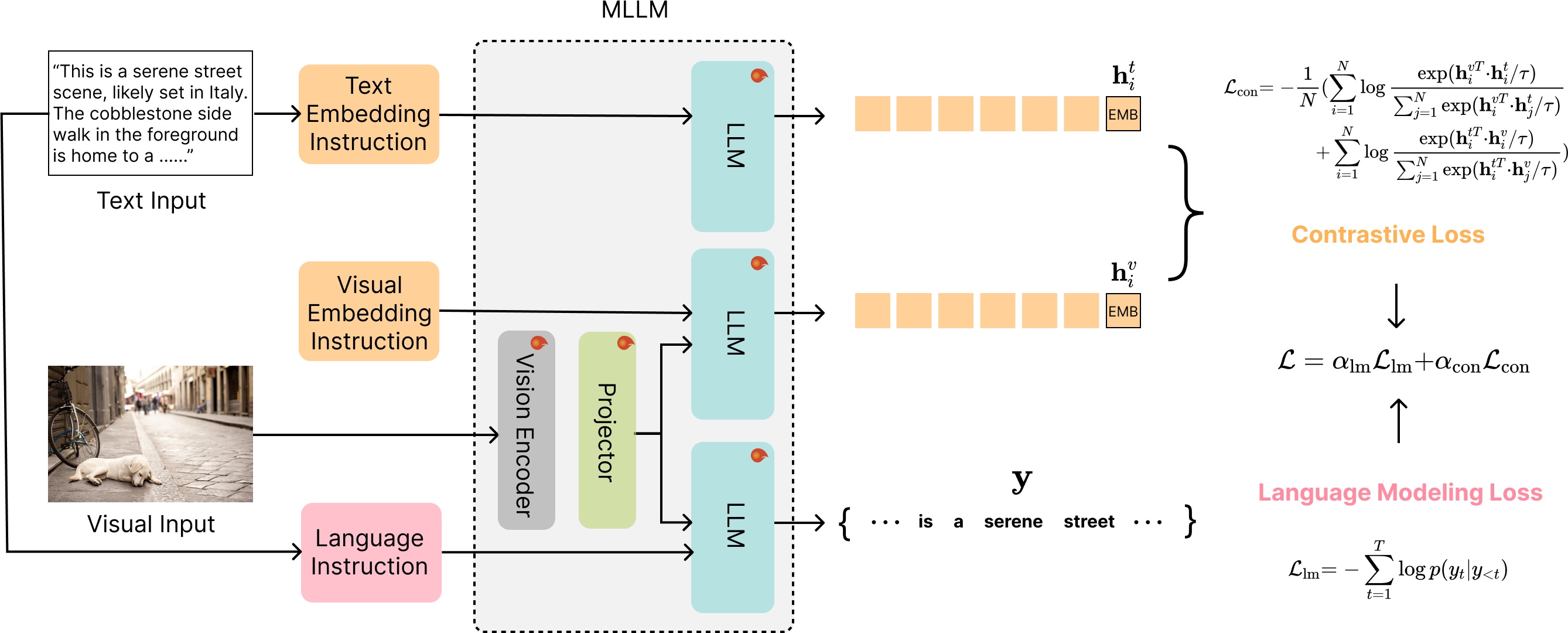}
    \caption{Pipeline of the proposed framework, \algname. It leverages a pretrained MLLM to jointly encode multimodal input and generate language responses. The model is trained using a weighted combination of contrastive loss and autoregressive language modeling loss on paired multimodal input. Specialized embedding instructions are designed to prompt the MLLM to generate effective embeddings, while language instructions are employed for language generation. The image is from MSCOCO dataset~\citep{Chen2015MicrosoftCC}. }
    \label{fig:pipeline}
\end{figure*}
We develop a unified framework for multimodal representation learning and natural language generation by adapting and jointly optimizing MLLM with both contrastive loss and language modeling loss. 
\figref{fig:pipeline} illustrates the pipeline of our proposed framework. 
We begin with a pretrained MLLM that processes multimodal inputs and generates language responses based on instructions. 
To enable embedding generation, we introduce an embedding instruction that mirrors the structure of language instructions, guiding the model to produce embeddings for any multimodal input. 
Fine-tuning is then conducted using paired image-text data, where the MLLM separately processes each modality to yield corresponding image and text embeddings. 
A contrastive loss is computed between these embeddings, while an autoregressive language modeling loss is applied via the language instruction of the same paired input. 
This joint fine-tuning enables the model to generate both high-quality free-form language responses and grounded robust multimodal embeddings.

\subsection{Multimodal embedding}\label{sec:emb_instruct}
We employ a prompt-based representation method to enable MLLMs to generate multimodal embeddings. 
Previous research has shown that prompt-based method can effectively represent sentence embedding with LLMs~\citep{jiang2023scaling}. 
Following this observation, we specifically design an embedding instruction to encode image and/or text, as shown in \tabref{tab:prompt}.

Following the conventional human-assistant conversation format of MLLMs~\citep{liu2024visual}, we set the system message and incorporate the image and text data as human input. 
Specifically, \texttt{<image>} represents the image token that encodes the visual information, while \texttt{<text>} encompasses any textual data that needs to be encoded.
Both \texttt{<image>} and \texttt{<text>} can be optionally absent, allowing for flexible encoding of multimodal or unimodal inputs (i.e., text-only or image-only). 
The generated assistant answer serves as the output, from which we extract the multimodal embedding by taking the last layer hidden states of the last token. 
Although the last token inherently captures all preceding information due to causal self-attention, following~\citet{jiang2023scaling}, we explicitly impose a one-word limitation, to enhance representation capabilities. 
Specifically, we use a \texttt{\{Embedding prompt\}} of ``\texttt{Compress this image/sentence in one word:}" for prompting the MLLM to generate embeddings of the multimodal input.
\begin{table}[t]
\centering
\begin{tcolorbox}[colback=yellow!20, colframe=black, arc=2mm]
\begin{verbatim}
System message
Human: <image> 
       <text>
       {Embedding prompt} 
Assistant: 
\end{verbatim}
\end{tcolorbox}
\caption{The embedding instruction used to generate multimodal embeddings with MLLMs. \texttt{<image>} and \texttt{<text>} represents the image token and text input, respectively. \texttt{\{Embedding prompt\}} is the prompt used for instructing the MLLM to generate the multimodal embedding. }
\label{tab:prompt}
\vspace{-0.2in}
\end{table}

\subsection{Contrastive-Autoregressive Joint Training}
We train the adapted MLLM jointly with a contrastive objective and autoregressive language modeling. 
Building on existing MLLM that has been pre-trained and instruction-tuned with autoregressive language modeling, we continue to fine-tune it using multimodal pairs for contrastive learning, while also maintaining its language generation capabilities through the autoregressive objective.

\paragraph{Contrastive learning.}
We employ the standard InfoNCE loss for contrastive learning, which encourages the model to learn discriminative representations that are close to positive samples and far from negative samples. 
The contrastive learning objective can be formulated as:
\begin{equation}
\begin{split}
    \mathcal{L}_{\text{con}} = - \frac{1}{N} (
    \sum_{i=1}^{N} \log \frac{\exp(\mathbf{h}_i^{v\intercal} \cdot \mathbf{h}_i^t / \tau)}{\sum_{j=1}^{N} \exp(\mathbf{h}_i^{v\intercal} \cdot \mathbf{h}_j^t / \tau)} \\ 
    + \sum_{i=1}^{N} \log \frac{\exp(\mathbf{h}_i^{t\intercal} \cdot \mathbf{h}_i^v / \tau)}{\sum_{j=1}^{N} \exp(\mathbf{h}_i^{t\intercal} \cdot \mathbf{h}_j^v / \tau)} 
    )
\end{split}
\label{eq:con}
\end{equation}
where $\mathbf{h_i^v}$ and $\mathbf{h_i^t}$ are the normalized embeddings of the $i$th visual-text pair, $N$ is the batch size, and $\tau$ is the temperature parameter that scale the logits. 

\paragraph{Autoregressive language modeling.} 
Following the standard approach described in~\citep{radford2018improving}, we train the model to predict the next token based on the context provided by the previous tokens. 
The autoregressive language modeling objective can be formulated as:
\begin{equation}
\mathcal{L}_{\text{lm}} = - \sum_{t=1}^{T} \log p(y_t | y_{<t})
\end{equation}
where $y_t$ is the target token at position t, and $p(y_t | y_{<t})$ is the predicted probability distribution over the vocabulary.

\paragraph{Contrastive-autoregressive modeling.} 
The overall objective function combines the autoregressive language modeling and contrastive learning objectives:
\begin{equation}
\mathcal{L} = \alpha_{\text{lm}} \mathcal{L}_{\text{lm}} + \alpha_{\text{con}} \mathcal{L}_{\text{con}} \label{eq:comb_loss}
\end{equation}
where $\alpha_{\text{lm}}$ and $\alpha_{\text{con}}$ are scaling parameters.

%% file: sec/4_experiments.tex
\section{Experiments}\label{sec:experiments}
\subsection{Training Setups}
\paragraph{Base models.}
We adopt LLaVa-OneVision~\citep{li2024llava} as the backbone MLLM for our joint contrastive-autoregressive training framework, given its state-of-the-art performance across various multimodal understanding tasks. 
However, note that our proposed \algname is model-agnostic, which is not limited to the choice of base model.

\paragraph{Datasets.}
LLaVa-OneVision~\citep{li2024llava} has been trained on a comprehensive and diverse instruction dataset that includes an important subset of detailed description data. 
This subset is particularly well-suited for contrastive learning, offering high-quality image-text pairs.
Specifically, we utilize subsets such as ShareGPT4V~\citep{chen2023sharegpt4v}, ShareGPT4o~\citep{sharegpt4o}, and Image Textualization~\citep{pi2024image}, comprising a total of 248K samples. 

One objective of this strategy is to demonstrate that, without introducing new data, MLLMs can generate high-quality vision-text embeddings by simply re-training on the existing dataset. 
And to adapt the instruction tuning data for contrastive learning, we modify the original samples by discarding their instructions and using the corresponding images as visual inputs.
The associated answers are then repurposed as captions to create image-text pairs.

\vspace{0.1cm}
\noindent\textbf{Implementation details.}
We fine-tune the pre-trained 0.5B and 7B single-image LLaVa-OneVision~\citep{li2024llava} models using a global batch size of 1024 and the combined contrastive and language modeling loss, as shown in \eqnref{eq:comb_loss}, for one epoch. 
To ensure that the contrastive loss and the language modeling loss are on a comparable scale, we set $\alpha_{\text{con}}$ to 10.0 and $\alpha_{\text{lm}}$ to 1.0 by default. 
Learning rates are set to $2\times10^{-6}$ for the vision encoder and $1\times10^{-5}$ for the projector and LLM. 
Other training configurations follow those of LLaVa-OneVision~\citep{li2024llava}. 
All experiments are conducted on 256 A100 GPUs.

\vspace{0.1cm}
\noindent\textbf{Evaluation strategies.}
We evaluate the model fine-tuned by \algname on both multimodal representation tasks and multimodal understanding tasks. 
The versatility of our framework allows it to seamlessly adapt to various representation and generation tasks without requiring additional fine-tuning or adaptation. 
Specifically, we present the results of our experiments on zero-shot image-text retrieval (\secref{subsec:img_text_retrieval}), multimodal retrieval (\secref{subsec:multimodal_retrieval}), multimodal understanding (\secref{subsec:multimodal_understanding}), and object hallucination reduction (\secref{subsec:oh_reduction}). 
%


\subsection{Zero-Shot Cross-modal Retrieval}\label{subsec:img_text_retrieval}
\begin{table*}[t]
\centering
\caption{
    Zero-shot image-text retrieval results on MSCOCO (5K test set)~\citep{Chen2015MicrosoftCC} and Flickr30K (1K test set)~\citep{flickrentitiesijcv} datasets. 
    Bold and underline indicate the first and second-best performance.
}
\setlength{\tabcolsep}{4pt}
\resizebox{\linewidth}{!}{
\begin{tabular}{lp{0.7cm}p{0.7cm}p{0.7cm}p{0.7cm}p{0.7cm}p{0.7cm}p{0.7cm}p{0.7cm}p{0.7cm}p{0.7cm}p{0.7cm}p{0.7cm}}
\toprule
& \multicolumn{6}{c}{Image $\rightarrow$ Text Retrieval} & \multicolumn{6}{c}{Text $\rightarrow$ Image Retrieval} \\
& \multicolumn{3}{c}{Flickr30K} & \multicolumn{3}{c}{MSCOCO} & \multicolumn{3}{c}{Flickr30K} & \multicolumn{3}{c}{MSCOCO} \\
&  R@1 & R@5 & R@10 & R@1 & R@5 & R@10 & R@1 & R@5 & R@10 & R@1 & R@5 & R@10 \\ 
\midrule
CLIP~\citep{radford2021learning}     & \textbf{88.0} & \textbf{98.7} & \textbf{99.4} & \underline{58.4} & \underline{81.5} & 88.1 & \underline{68.7} & \underline{90.6} & \underline{95.2} & 37.8 & 62.4 & 72.2 \\
VLM2VEC~\citep{jiang2024vlm2vec}     & 84.3 & 96.6 & 98.8 & 58.2 & 81.4 & \underline{88.9} & 66.7 & 88.8 & 93.9 & \underline{38.8} & \underline{66.7} & \underline{77.5} \\
LLaVA-OV-SI-0.5B~\citep{li2024llava} & 50.6 & 77.5 & 85.7 & 25.1 & 49.0 & 60.8 & 39.4 & 67.9 & 78.0 & 20.7 & 42.6 & 54.1 \\
LLaVA-OV-SI-7B~\citep{li2024llava} & 31.4 &	59.2 &	71.3 &	14.5 &	33.5 &	44.6 &	46.3 &	72.9 &	80.6 &	18.2 &	38.5 &	48.9\\
\midrule
\algname-0.5B & 80.5 & 95.3 & 97.6 & 57.2 & 81.3 & 88.5 & 66.1 & 89.2 & 93.3 & 39.2 & 66.2 & 76.1 \\
\algname-7B & \underline{87.5} & \underline{98.2} & \underline{99.2} & \textbf{64.6} & \textbf{85.9} & \textbf{91.8} & \textbf{75.3} & \textbf{92.6} & \textbf{96.0} & \textbf{47.9} & \textbf{73.7} & \textbf{82.4} \\
\bottomrule
\end{tabular}
}
\label{tab:zero_ret}
\end{table*}
Model finetuned by \algname can be directly applied to image-text retrieval tasks, as it can generate aligned image and text unimodal embeddings. 
To evaluate its performance, we conduct experiments on two standard image-text retrieval benchmarks: MSCOCO (5K test set)~\citep{Chen2015MicrosoftCC} and Flickr30K (1K test set)~\citep{flickrentitiesijcv} in the zero-shot setting. 
In this setting, all parameters are frozen and the model is directly used to extract embeddings without any fine-tuning. 
Following the CLIP~\citep{radford2021learning}, we first generate embeddings for each image and text sample independently for all image-text pairs in the test set. 
We then compute cosine similarity scores between these embeddings and use them to retrieve relevant image-text pairs from the entire test set. 
We report the recall for the top 1, top 5, and top 10 retrieved candidates for both image to text retrieval and text to image retrieval in \tabref{tab:zero_ret}. 

We compare model finetuned by \algname with several strong baselines. 
VLM2VEC~\citep{jiang2024vlm2vec} is the state-of-the-art model that also adapts a MLLM for multimodal embeddings. 
We take their model checkpoint released on HuggingFace\footnote{https://huggingface.co/TIGER-Lab/VLM2Vec-LoRA}. 
LLaVa-OneVision~\citep{li2024llava} is the MLLM that our model builds on and we report results for both 0.5B and 7B model (LLaVA-OV-SI-0.5B and LLaVA-OV-SI-7B). 
We use the same embedding instruction as described in \secref{sec:emb_instruct} and directly extract the last layer hidden states of the last token as the image and text embeddings. 
This baseline shows the zero-shot embedding capabilities of a MLLM that has not been finetuned using our approach. 
Last but not least, we include CLIP~\citep{radford2021learning}, which is a pioneering vision-language foundation model that achieves impressive zero-shot performance on image-text retrieval, as our baseline as well. 

Results in \tabref{tab:zero_ret} show that by finetuning LLaVa-OneVision model using the proposed contrastive-autoregressive loss, our \algname-7B model outperforms all the baselines. 
The 0.5B model, which is much smaller, also exhibits strong performance. 
Note that the original zero-shot LLaVA-OV-SI-0.5B shows relatively worse performance, indicating that MLLM embeddings are not ideal for representation learning and require contrastive fine-tuning to achieve optimal results. 
Additionally, the performance of the 7B model LLaVA-OV-SI-7B does not surpass that of the 0.5B model LLaVA-OV-SI-0.5B, which may be due to the model not being optimized for effective representation, indicating that a larger MLLM does not necessarily lead to enhanced representation capabilities.

\subsection{Multimodal Retrieval}\label{subsec:multimodal_retrieval}
\begin{table*}[t]
\centering
\caption{Multimodal retrieval results on the Multimodal Embedding Benchmark (MMEB), with scores averaged across each meta-task, as well as average scores for in-distribution datasets (IND), out-of-distribution datasets (OOD) and all datasets (Overall).}
\setlength{\tabcolsep}{6pt}
\resizebox{\linewidth}{!}{
\begin{tabular}{cccccccc}
\toprule
Model & Classification & VQA & Retrieval & Grounding & IND & OOD & Overall \\
\midrule
CLIP~\citep{radford2021learning} & 42.8 & 9.1 & 53.0 & 51.8 & 37.1 & 38.7 & 37.8 \\
BLIP2~\citep{li2023blip} & 27.0 & 4.2 & 33.9 & 47.0 & 25.3 & 25.1 & 25.2\\
SigLIP~\citep{zhai2023sigmoid} & 40.3 & 8.4 & 31.6 & 59.5 & 32.3 & 38.0 & 34.8\\
OpenCLIP~\citep{cherti2023reproducible} & 47.8 & 10.9 & 52.3 & 53.3 & 39.3 & 40.2 & 39.7\\
UniIR~\citep{wei2023uniir} & 42.1 & 15.0 & 60.1 & 62.2 & 44.7 & 40.4 & 42.8\\
E5-V~\citep{jiang2024e5} & 21.8 & 4.9 & 11.5 & 19.0 & 14.9 & 11.5 & 13.3 \\
Magiclens~\citep{zhang2024magiclens} & 38.8 & 8.3 & 35.4 & 26.0 & 31.0 & 23.7 & 27.8\\
VLM2VEC~\citep{jiang2024vlm2vec} & 54.8 & \underline{54.9} & \underline{62.3} & 79.5 & \underline{66.5} & 52.0 & \underline{60.1}\\
LLaVA-OV-SI-7B~\citep{li2024llava} & 19.8 & 6.9 & 17.3 & 16.0 & 17.0 & 12.4 & 14.9 \\
\midrule
\algname-0.5B & \underline{59.1} & 49.1 & 61.0 & \underline{83.0} & 64.3 & \underline{53.7} & 59.6 \\
\algname-7B & \textbf{65.2} & \textbf{65.6} & \textbf{70.0} & \textbf{91.2} & \textbf{75.8} & \textbf{62.4} & \textbf{69.8} \\
\bottomrule
\end{tabular}
}
\label{tab:vlm}
\end{table*}
Multimodal retrieval involves queries and targets that include interleaved text and images. 
It is more challenging than the traditional text-image retrieval tasks, as the model needs to not only understand the individual modalities but also capture their complex relationships and interplay. 
Traditional methods~\citep{radford2021learning,li2023blip,zhai2023sigmoid,cherti2023reproducible} often fall short by either processing text and images separately or performing shallow fusion of visual and textual information. 
In contrast, MLLMs offer an ideal solution, leveraging the multimodal reasoning and understanding abilities of LLMs to effectively integrate and process multiple modalities.
We evaluate our model on Massive Multimodal Embedding Benchmark (MMEB)~\citep{jiang2024vlm2vec}, which includes 36 datasets spanning four meta-task categories: classification, visual question answering, retrieval, and visual grounding. 
For fair comparison, we finetuned our model on the training sets of MMEB using contrastive loss only for 600 steps. 
Results on the validation set of MMEB are reported in \tabref{tab:vlm}. 
Following~\cite{jiang2024vlm2vec}, Precision@1 scores are averaged for each meta-task and we also provide an average score for in-distribution datasets (IND), out-of-distribution datasets (OOD), as well as an overall average score across all datasets. 

For comparison, we include several baselines that directly apply vision/text encoders (CLIP~\citep{radford2021learning}, BLIP2~\citep{li2023blip}, SigLIP~\citep{zhai2023sigmoid}, OpenCLIP~\citep{cherti2023reproducible}) and combine multimodal features using score-level fusion by element-wise addition with equal weights following~\cite{jiang2024vlm2vec}. 
We also include UniIR~\citep{wei2023uniir}, which builds on CLIP and BLIP, employing shallow fusion techniques such as score-level and feature-level fusion to integrate modalities. 
MagicLens~\citep{zhang2024magiclens} uses a multi-head attention pooler to unify multimodal inputs into a single embedding. 
E5-V~\citep{jiang2024e5} leverages vision-language models for multimodal embedding tasks and is trained exclusively on text pairs. 
VLM2REC~\citep{jiang2024vlm2vec} is the current state-of-the-art model on this benchmark, which also leverages vision-language models for multimodal embedding tasks. 
Additionally, we include LLaVA-OV-SI-7B~\citep{li2024llava}, which extracts embeddings from LLaVA-OV-SI-7B directly, demonstrating the zero-shot embedding capabilities of an MLLM that has not been fine-tuned with contrastive loss, similar to the one we have in cross-modal retrieval (\secref{subsec:img_text_retrieval}).

The results in \tabref{tab:vlm} demonstrate that our method significantly outperforms all baselines and the previous state-of-the-art by a large margin. 
Notably, our 0.5B model, which is 8 times smaller than VLM2REC~\citep{jiang2024vlm2vec}, achieves comparable performance. 
Our 7B model improves upon the previous state-of-the-art by a 10.2 percentage points in overall score, indicating the effectiveness of \algname.

\subsection{Multimodal Understanding}\label{subsec:multimodal_understanding}

\begin{table}[t]
\centering
\caption{
    Results on MMMU and MMStar. 
    \algname performs comparably with LLaVA-OV~\citep{li2024llava} on multimodal understanding.
}
\begin{tabular}{lccc}
\toprule
Model & MMMU & MMStar\\
\midrule
LLaVA-OV-SI-0.5B~\citep{li2024llava} & 31.2 & 36.3 \\
\algname-0.5B & 31.0 & 39.1 \\
\midrule
LLaVA-OV-SI-7B~\citep{li2024llava} & 47.3 & 60.9 \\
\algname-7B & 47.0 & 56.8 \\
\bottomrule
\end{tabular}
\label{tab:understanding}
\end{table}

We also evaluate our method on multimodal understanding and reasoning in zero-shot manner, focusing on widely-adopted benchmarks including 
MMMU~\citep{yue2024mmmu} and MMStar~\citep{chen2024we}.
MMMU~\citep{yue2024mmmu} is a benchmark that includes 11.5K meticulously collected multimodal, multi-discipline questions from college exams, quizzes, and textbooks, requiring college-level subject knowledge and deliberate reasoning. MMStar~\citep{chen2024we} comprises 1,500 multimodal, multi-discipline questions meticulously selected by humans for visual dependency and minimal data leakage, demanding advanced multimodal capabilities. 
%
We compare against the original LLaVa-OneVision~\citep{li2024llava} to assess the impact of contrastive-autoregressive fine-tuning on the multimodal understanding capabilities of the MLLM. Overall, we observe comparable performance to the state-of-the-art MLLM, LLaVa-OneVision on these datasets. 

\subsection{Object Hallucination Reduction}\label{subsec:oh_reduction}

\begin{table*}[t]
\centering
\caption{
    THRONE Results on MSCOCO validation set. Class-wise and overall scores are reported. 
    \algname outperforms baselines and achieves state-of-the-art performance on mitigating object hallucinations.
}
\resizebox{\linewidth}{!}{
\begin{tabular}{ccccccccccc}
\toprule
Model & $P_{ALL}$ & $R_{ALL}$ & $F^1_{ALL}$ & $F^{0.5}_{ALL}$ & $P_{CLS}$ & $R_{CLS}$ & $F^1_{CLS}$ & $F^{0.5}_{CLS}$ \\
\midrule
Adapter-v2~\citep{gao2023llama} & 63.6 & 73.3 & 68.1 & 65.3 & 68.2 & 70.6 & 69.4 & 68.7 \\
Adapter-v2.1~\citep{gao2023llama} & 63.8 & 73.7 & 68.4 & 65.5 & 67.4 & 71.2 & 69.3 & 68.1 \\
InstructBLIP~\citep{dai2023instructblip} & 70.8 & 74.3 & 72.5 & 71.5 & 77.2 & 71.9 & 74.5 & 76.1 \\
Otter-Image~\citep{li2023otter} & 33.0 & 31.2 & 32.1 & 32.7 & 25.2 & 16.9 & 20.2 & 22.9 \\
MiniGPT4~\citep{zhu2023minigpt} & 81.7 & 59.8 & 69.0 & 76.1 & 79.9 & 61.8 & 69.7 & 75.5 \\
MiniGPT-v2~\citep{chen2023minigpt} & 79.0 & 66.6 & 72.3 & 76.2 & 77.6 & 67.0 & 71.9 & 75.2 \\
mPLUG-Owl~\citep{ye2023mplug}  & 55.5 & 71.9 & 62.6 & 58.1 & 66.3 & 68.3 & 67.3 & 66.7 \\
LRV-Instruction-v2~\citep{liu2023aligning} & 82.0 & 56.7 & 67.0 & 75.3 & 78.4 & 58.8 & 67.2 & 73.5 \\
LLaVA-v1.3~\citep{liu2024visual} & 80.5 & 65.2 & 72.1 & 76.9 & 79.9 & 65.3 & 71.9 & 76.5 \\
LLaVA-v1.5~\citep{liu2024improved} & 68.1 & 61.0 & 64.4 & 66.6 & 69.9 & 56.4 & 62.5 & 66.8 \\
LLaVA-Mistral~\citep{jiang2023mistral,liu2024llava} & 86.8 & 71.8 & 78.3 & 83.6 & 84.4 & 64.2 & 70.8 & 77.5 \\
LLaVA-OV-SI-7B~\citep{li2024llava}& 86.8 & \textbf{81.8} & 85.0 & 87.3 & 86.9 & 77.1 & 79.6 & 83.3 \\
\midrule
\algname & \textbf{87.0} & 81.5 & \textbf{86.3} & \textbf{88.6} & \textbf{88.1} & \textbf{78.2} & \textbf{81.2} & \textbf{84.8} \\
\bottomrule
\end{tabular}
}
\label{tab:throne_coco}
\end{table*}

\begin{table}[t]
\centering
\caption{POPO results on MSCOCO validation set. Yes denotes
the proportion of answering “Yes” to the given question. \algname achieves the highest F1 score and accuracy.}
\setlength{\tabcolsep}{3pt}
\begin{tabular}{cccc|c|c}
\toprule
Model & Acc. & Prec. & Recall & F1 & Yes (\%) \\
\midrule
mPLUG-Owl~\citep{ye2023mplug} & 51.5 & 50.8 & 99.4 & 67.2 & 97.8 \\
LLaVA~\citep{liu2024visual} & 52.5 & 51.3 & 99.8 & 67.8 & 97.3 \\
MultiModal-GPT~\citep{gong2023multimodal} & 50.0 & 50.0 & \textbf{100.0} & 66.7 & 100.0 \\
MiniGPT-4~\citep{zhu2023minigpt} & 70.9 & 67.4 & 82.8 & 74.1 & 61.9 \\
InstructBLIP~\citep{dai2023instructblip} & 81.5 & 75.9 & 93.7 & 83.7 & 62.2 \\
LLaVA-OV-SI-7B~\citep{li2024llava} & 89.0 & \textbf{95.9} & 81.4 & 88.1 & 42.4 \\
\midrule
\algname & \textbf{89.1} & 91.0 & 87.0 & \textbf{88.9} & 47.9 \\
\bottomrule
\end{tabular}
\label{tab:pope}
\end{table}

Hallucination in LLMs and MLLMs has recently attracted significant research attention~\citep{ye2023cognitive,zhang2023siren,biten2022let,zhou2023analyzing, chen2024halc, fang2024uncertainty}. MLLMs have been shown to generate text that contradicts the visual or textual input, specifically by producing nonexistent objects in images (object hallucination)~\citep{biten2022let,zhou2023analyzing}. Such hallucinations can degrade model performance and severely impact user experience in real-world applications. Consequently, we also evaluate our model's performance on object hallucinations with THONE~\citep{kaul2024throne} and POPE~\citep{li2023evaluating}. 

\noindent\textbf{Results on THRONE.} THONE employs language models to assess object hallucinations in free-form, open-ended image descriptions with respect to a predefined vocabulary of objects of interest. It utilizes the validation set of COCO 2017~\citep{Chen2015MicrosoftCC}, which comprises 5,000 images and 80 annotated object categories, prompting MLLMs to generate descriptive responses for each image. Subsequently, FLAN-T5 models~\citep{longpre2023flan} are used to evaluate whether the MLLMs produce descriptions containing nonexistent objects. 

Results are presented in \tabref{tab:throne_coco}, where popular MLLMs with $\sim$7B parameters are compared. Overall and class-wise precision, recall, F0.5-score, F1-score are reported and following~\cite{kaul2024throne} classwise F0.5-score is considered the principal metric for evaluating model performance. As shown in \tabref{tab:throne_coco}, while LLaVA-OV-SI-7B~\citep{li2024llava} achieves the best performance among all the MLLMs listed, our proposed method with contrastive and autoregressive finetuning improves further upon LLaVA-OV-SI-7B for 1.5 pp of classwise F0.5-score. The performance gain could be likely due to the contrastive training from which the model learns to align representations of images and text. By learning to distinguish between matched and mismatched pairs, the model could potentially develop a more accurate understanding of visual and textual correspondences, and consequently reduce the likelihood of generating hallucinated objects that don't match the context.

\noindent\textbf{Results on POPE.} The Polling-based Object Probing Evaluation (POPE)~\citep{li2023evaluating} is a method for assessing object hallucination in LVLMs. It involves asking LVLMs questions like "Is there a [object] in the image?", with responses alternating between "Yes" and "No," ensuring am equal 50\% probability for each answer. POPE is divided into three splits: random, popular, and adversarial. The random split involves randomly chosen missing objects, the popular split focuses on frequently occurring objects, and the adversarial split targets objects closely related to those in the image. The evaluation uses 500 randomly selected images from the MSCOCO~\citep{Chen2015MicrosoftCC} validation set. The scores, including precision, recall, F1 score, accuracy, and yes ratio, are averaged across the three splits and presented in \tabref{tab:pope}. The proposed \algname consistently outperforms all the baselines in accuracy and F1 score, indicating the model's effectiveness in mitigating object hallucinations.

%% file: sec/5_ablation_analysis.tex
\section{Ablation and Analysis}
\begin{figure*}[t]
    \centering
    \hspace{0.5in}MSCOCO~\citep{Chen2015MicrosoftCC} \hspace{1in} Flickr30K~\citep{flickrentitiesijcv} \\\includegraphics[width=0.248\textwidth]{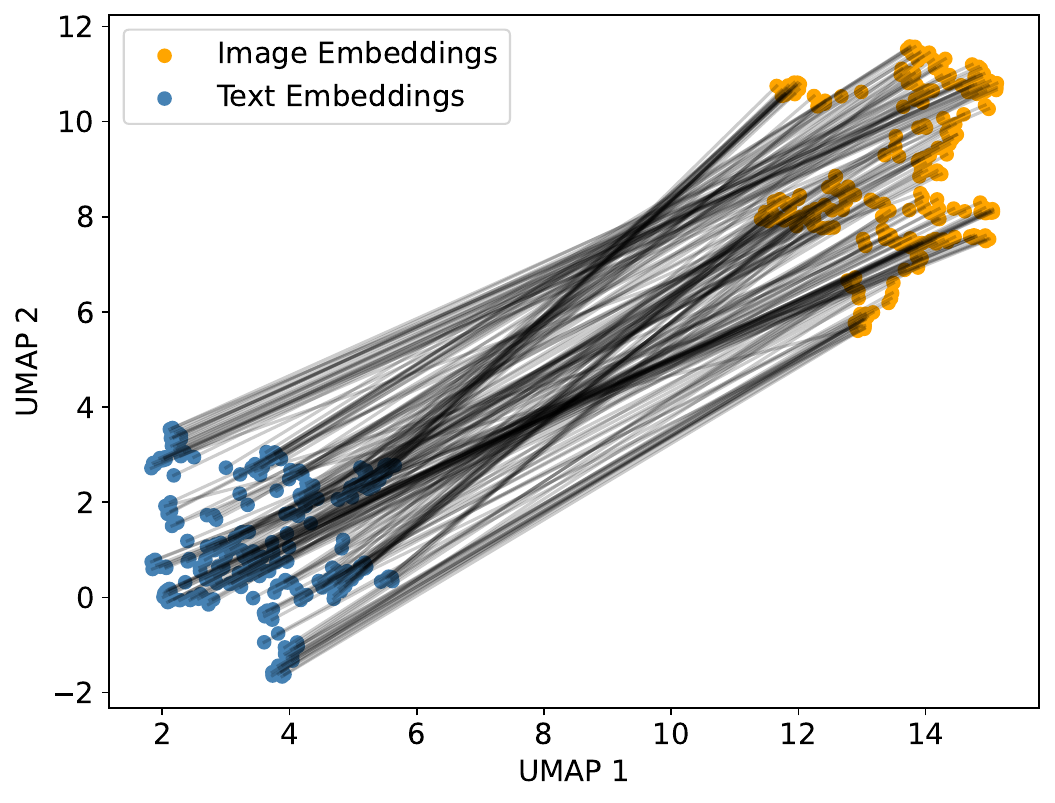}
    \includegraphics[width=0.243\textwidth]{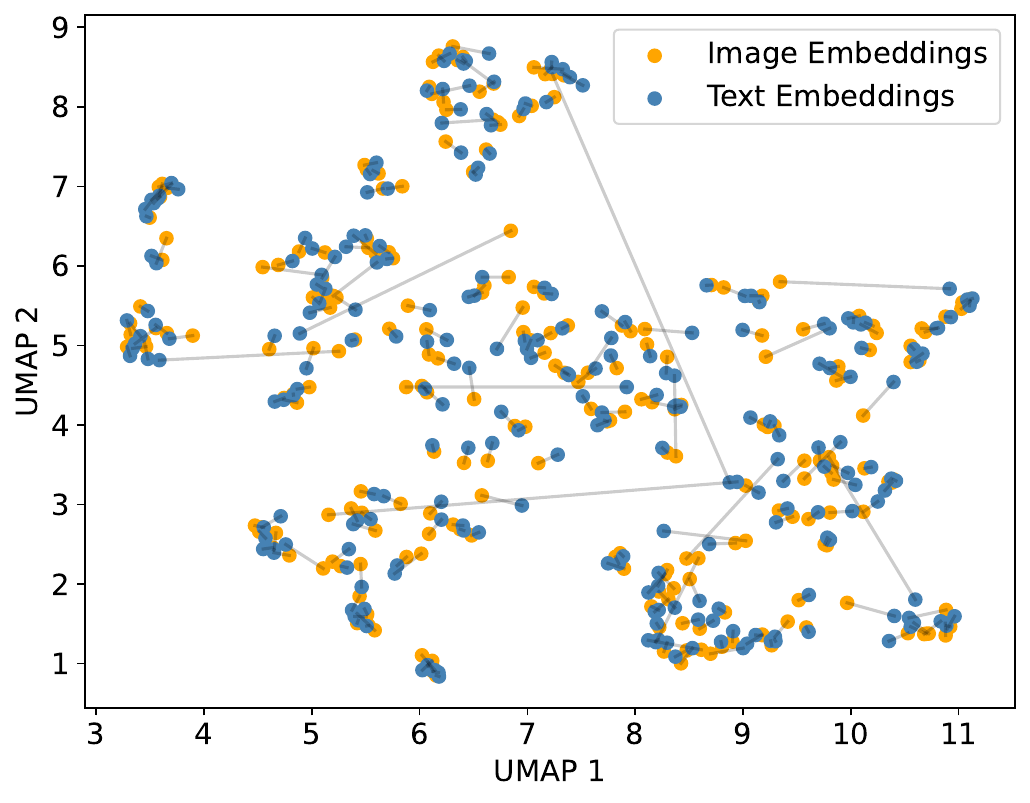}  
    \includegraphics[width=0.243\textwidth]{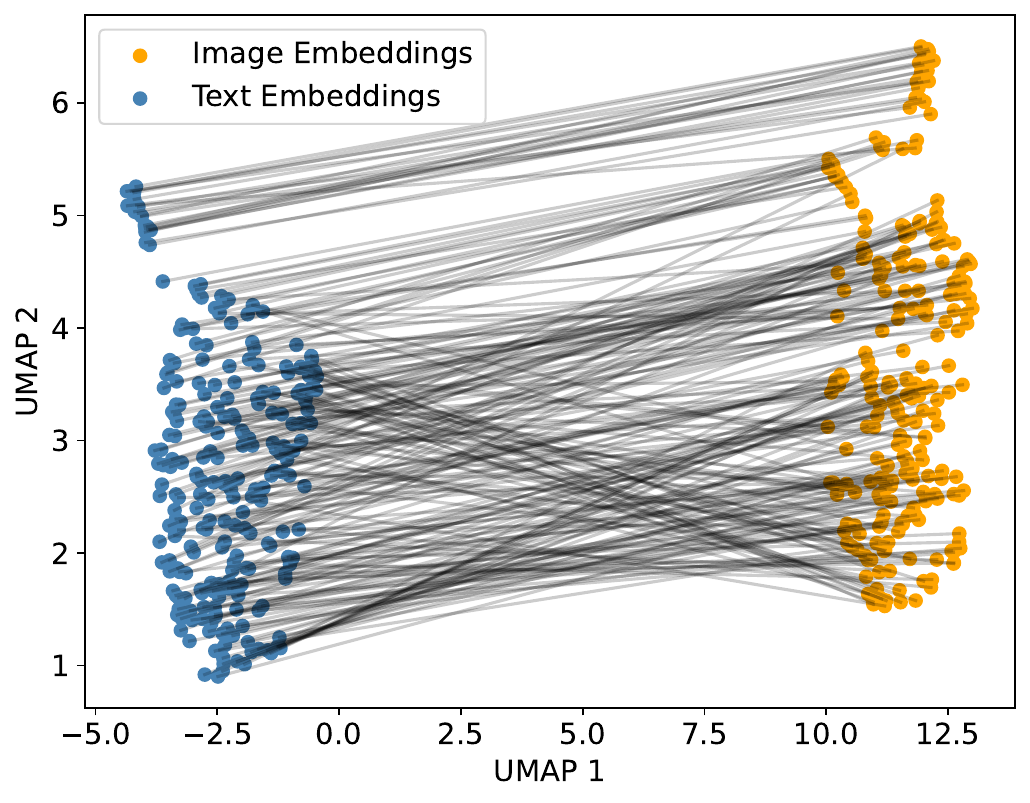}
    \includegraphics[width=0.243\textwidth]{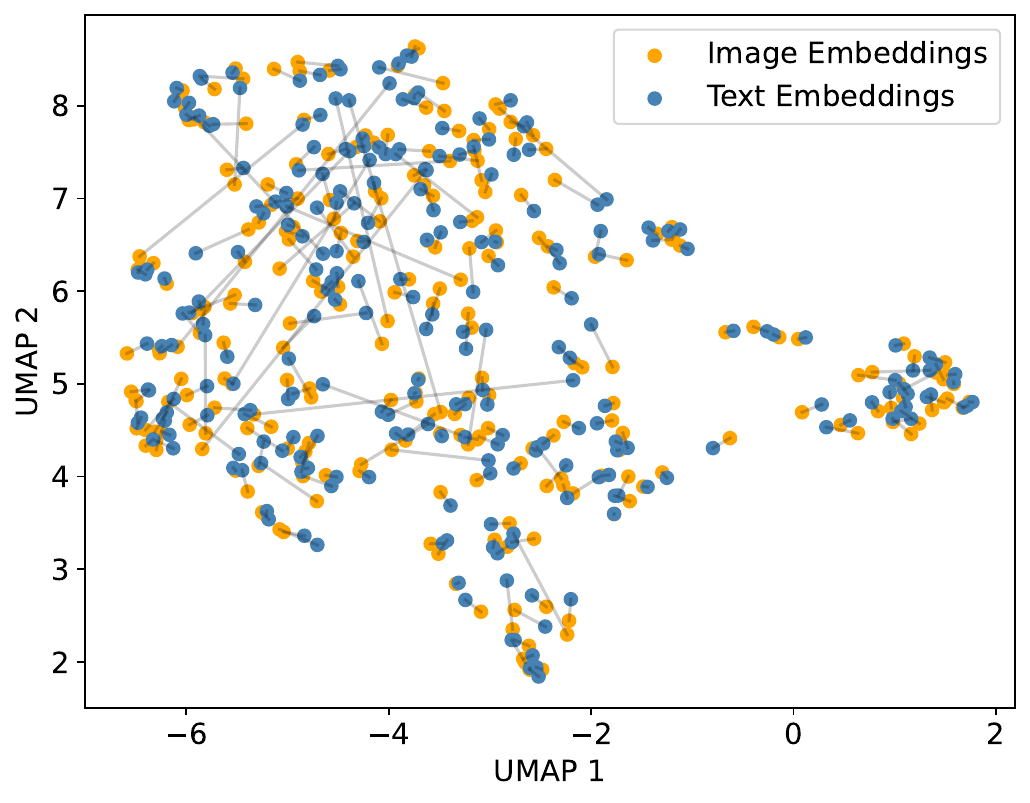}
    LLaVA-OV~\citep{li2024llava}\hspace{0.5in}\algname \hspace{0.6in} LLaVA-OV~\citep{li2024llava} \hspace{0.4in} \algname \hspace{0.4in}
    
    \caption{Scatter plots of image-text embeddings from MSCOCO~\citep{Chen2015MicrosoftCC} and Flickr30K~\citep{flickrentitiesijcv} datasets. \algname removes the existing modality gap in multimodal representation space.}
    \label{fig:gap}
\end{figure*}
\subsection{Ablation on Training Objectives}
\begin{table}[t]
\centering
\caption{
    Performance comparison across varying weights for language modeling ($\alpha_{\text{lm}}$) and contrastive ($\alpha_{\text{con}}$) objectives.
    Zero-shot retrieval recall@1 is reported for image-to-text (I2T) and text-to-image (T2I) tasks on the Flickr and MSCOCO datasets, alongside multimodal understanding scores on MMMU and MMStar. 
    A ratio of 0 indicates training solely with contrastive loss, while a ratio of 1 implies equal emphasis on both objectives.
}
\setlength{\tabcolsep}{4pt}
\begin{tabular}{ccccccc}
\toprule
\multicolumn{1}{c}{$\alpha_{\text{lm}}:\alpha_{\text{con}}$} 
& \multicolumn{2}{c}{Flickr} 
& \multicolumn{2}{c}{MSCOCO} 
& MMMU & MMStar \\
\cmidrule(lr){2-3} \cmidrule(lr){4-5}
& I2T & T2I & I2T & T2I & & \\
\midrule
0 & 89.0 & 75.0 & 67.0 & 49.6 & 44.4 & 53.9 \\
0.05 & 88.3 & 75.8 & 65.3 & 48.6 & 47.0 & 56.5 \\
0.1 & 87.5 & 75.3 & 64.4 & 47.9 & 47.0 & 56.8 \\
0.2 & 87.0 & 74.7 & 63.3 & 47.0 & 47.4 & 57.3 \\
1 & 84.9 & 73.2 & 61.1 & 45.1 & 47.4 & 57.1 \\
\bottomrule
\end{tabular}
\label{tab:loss}
\end{table}
%

We study the effects of two training objectives and compare different variants of the model in \figref{tab:loss}. Given that the scale of the contrastive loss is one-tenth that of the language modeling loss, we experimented with different weightings for the losses, specifically using ratios of  $\alpha_{\text{lm}}:\alpha_{\text{con}}$ = 0, 0.02, 0.1, 0.2, and 1. Among them, $\alpha_{\text{lm}}:\alpha_{\text{con}}$ = 0 is the case where the model is trained using only the contrastive loss and $\alpha_{\text{lm}} = 0$. Results indicate that generally a larger contrastive loss leads to better representation abilities but worse performance on multimodal understanding tasks. We present a trade-off between representation abilities and generation abilities.  

\subsection{Modality Gap}

Modality gap~\citep{liang2022mind} refers to a geometric phenomenon in the representation spaces of multimodal models, where different data modalities, such as images and texts, are embedded at a significant distance from each other, rather than being uniformly distributed as ideally expected. 
This gap, which originates from initialization and persists throughout the contrastive learning process (as seen in models like CLIP~\citep{radford2021learning}), presents a challenge for language-image pretraining by affecting joint data modeling and understanding. Recent studies~\citep{srivastava2024omnivec,oh2024geodesic} suggest that narrowing this gap could improve multimodal representations and enhance performance in downstream tasks. 

We analyze the modality gaps in the multimodal embedding space of the proposed \algname by visualizing the embeddings of 250 randomly selected image-text pairs in MSCOCO and Flickr datasets. Following~\citep{liang2022mind}, we adopt UMAP~\citep{mcinnes2018umap} to reduce the dimensionality of the embeddings to 2 and  
compare the embeddings of the proposed method with LLaVA-OV~\citep{li2024llava} in \figref{fig:gap}.
As shown in the figure, even though LLaVA-OV~\citep{li2024llava} generated both text and image embeddings using the same model, there exists a significant modality gap between text and images. 
In contrast, after finetuning with contrastive-autoregressvie loss, the proposed model, \algname, removes this modality gap in the generated embeddings, where embeddings are aligned based on semantic meaning rather than modality.

%% file: sec/6_conclusion.tex
\section{Conclusion}

We introduce \algname, a contrastive-autoregressive fine-tuning framework designed to enhance MLLMs for both representation and generative tasks. Our approach adapts a pretrained MLLM to generate multimodal embeddings while maintaining its language generation capabilities by designing novel embedding instructions. We further propose to fine-tune the model by integrating a contrastive objective with autoregressive language modeling loss.

We conducted extensive experiments on retrieval for both cross-modal (image-text) and multimodal tasks. Our results demonstrate significant performance improvements, particularly in multimodal retrieval tasks. This is likely due to the model's ability to leverage the strengths of LLMs in processing and understanding multimodal information, surpassing traditional multimodal fusion methods such as simple concatenation. Additionally, we also demonstrate that the proposed framework removes the modality gap in the multimodal representation space, potentially improving cross-modal consistency and enhancing multimodal performance.
Furthermore, our proposed framework also demonstrates strong multimodal understanding capability and robustness against object hallucinations. Results on THRONE and POPE highlight the superior performance in mitigating object hallucinations. 
Overall, we present a multimodal framework that effectively and seamlessly combines representation learning and generation capabilities, providing insights for developing better-rounded multimodal models.
%


%% file: sec/X_appendix.tex
\section{Qualitative Case Study on Hallucination Reduction}\label{app:case_study}
We present a qualitative case study on object hallucination using the MSCOCO dataset~\citep{Chen2015MicrosoftCC}. Captions generated using LLaVA-OV~\citep{li2024llava} and using our proposed \algname are compared. Text in red indicates objects that are not present in the original image but are generated by the model. According to the case study, LLaVA-OV~\citep{li2024llava} generated hallucinated objects, \textit{e.g.}, 'sink', 'bird'. Our proposed \algname demonstrates superior performance by generating fewer object hallucinations. Additionally, it produces longer and more detailed captions compared to LLaVA-OV~\citep{li2024llava}.

\begin{tcolorbox}[colback=blue!5!white, colframe=blue!50!black, width=\textwidth, sharp corners, boxrule=0.5mm]
        \centering 
        \includegraphics[width=0.35\linewidth]{fig/000000191013.jpg} 
        \vspace{0.5em}
    \noindent
    \begin{tikzpicture}
        \draw[dashed] (0,0) -- (\textwidth,0); 
    \end{tikzpicture}
    \vspace{0.5em}
        \parbox[c]{\linewidth}{ 
            \textbf{\textit{LLaVA-OV~\citep{li2024llava} Generated Caption:}} In the heart of a bustling city, a large, vibrant mural adorns the side of a brick building. The mural is a lively depiction of a yellow \textcolor{green}{clock} with black hands and numbers, standing out against the red background. The \textcolor{green}{clock} is not alone; it's accompanied by a red \textcolor{red}{bird} perched on its top and a gold trumpet resting on its bottom, adding an element of whimsy to the scene.
            The building itself is situated on a busy street corner. A yellow taxi cab and a red van are parked in front of it, hinting at the hustle and bustle of city life. Above, the sky is a clear blue, dotted with white clouds, providing a serene backdrop to this urban landscape.
            The image captures the essence of city life - the blend of art and architecture, the constant movement of vehicles, and the ever-changing sky. It's a snapshot of a moment in time in a city that's always on the move.
            \\
            \textcolor{blue}{\textbf{\algname (Ours) Generated caption:}} The image captures a large, ornate \textcolor{green}{clock} mounted on the side of a building. The \textcolor{green}{clock} face is circular and prominently features Roman numerals for the hours, with the hands indicating a time of approximately 12:30. Above the \textcolor{green}{clock} face, there is a decorative element that includes a red shield with the Vodafone logo in white, flanked by two golden trumpets on either side. Below the \textcolor{green}{clock} face, there are two banners; the upper banner reads "İşte ben de bir kral olur" which translates to "I too want to be a king" in Turkish, while the lower banner states "Vodafone 24 saat hizmet" meaning "Vodafone 24-hour service."
            In the foreground, there are several vehicles parked along the street, including a yellow taxi cab, a white van, and a maroon SUV. The sky above is clear with a few clouds, suggesting it might be late afternoon or early evening based on the lighting.
            The building itself appears to be constructed with red brick, and there is another advertisement visible on the right side of the building, though it is partially obscured by the \textcolor{green}{clock} structure. The overall style of the image is a real-life photograph taken during daylight hours.
        }
\end{tcolorbox}

\begin{tcolorbox}[colback=blue!5!white, colframe=blue!50!black, width=\textwidth, sharp corners, boxrule=0.5mm]
        \centering 
        \includegraphics[width=0.4\linewidth]{fig/000000237928.jpg} 
        \vspace{0.5em}
    \noindent
    \begin{tikzpicture}
        \draw[dashed] (0,0) -- (\textwidth,0); 
    \end{tikzpicture}
    \vspace{0.5em}
        \parbox[c]{\linewidth}{ 
            \textbf{\textit{LLaVA-OV~\citep{li2024llava} Generated Caption:}} This image captures a modern kitchen bathed in soft light. The dominant color scheme is white, punctuated by the sleek silver of the appliances and the vibrant red of the kettle. 
            The kitchen is equipped with a stainless steel \textcolor{green}{refrigerator} standing tall on the right, its surface reflecting the ambient light. Adjacent to it, a black stove waits ready for culinary adventures, while a \textcolor{green}{microwave} sits atop it, offering convenience at arm's reach.
            The heart of the kitchen is a \textcolor{red}{sink} nestled between the stove and a window. The window allows natural light to filter in, illuminating the space and creating a warm atmosphere. Above the sink, white cabinets offer ample storage space, their pristine surfaces adding to the overall clean aesthetic of the kitchen.
            A red kettle rests on the counter next to the sink, its bold color contrasting beautifully with the surrounding whites. Its presence suggests many mornings started with a comforting cup of tea.
            Overall, this kitchen exudes a sense of modern elegance and functionality, with every element thoughtfully placed for ease of use and visual appeal.
            \\
            \textcolor{blue}{\textbf{\algname (Ours) Generated caption:}} The image depicts a modern kitchen with a clean and organized appearance. The kitchen features white cabinetry that includes both upper and lower cabinets, providing ample storage space. The upper cabinets have two doors each, while the lower cabinets have one door each, all fitted with silver handles.
            On the countertop, there is a stainless steel \textcolor{green}{microwave} \textcolor{green}{oven} positioned to the left of a red kettle. The \textcolor{green}{microwave} has a digital display and control panel on its front side. Next to the \textcolor{green}{microwave}, there is a red kettle with a black handle, which adds a pop of color to the otherwise neutral-toned kitchen.
            To the right of the \textcolor{green}{microwave} and kettle, there is a stainless steel \textcolor{green}{refrigerator} with a freezer compartment at the bottom. The \textcolor{green}{refrigerator} has a sleek design with a handle located on the right side for accessing the freezer and a separate handle for the main compartment.
            Below the countertop, there is a built-in \textcolor{green}{oven} with a glass window and a red \textcolor{green}{oven} mitt hanging from the \textcolor{green}{oven} handle. The \textcolor{green}{oven} appears to be part of the cabinetry, seamlessly integrated into the design.
            The backsplash behind the countertop consists of dark red tiles, which contrast with the white cabinetry and add depth to the space. The floor is covered with light-colored carpeting, which complements the overall color scheme of the kitchen.
            The lighting in the kitchen is soft and ambient, coming from recessed lights in the ceiling, which creates a warm and inviting atmosphere. There are no visible windows or other sources of natural light in the image.
            Overall, the kitchen is designed with functionality and aesthetics in mind, featuring modern appliances and a cohesive color palette that combines white, silver, and red accents.
            \\  
        }
    
    
\end{tcolorbox}


